\documentclass[12pt]{article}
\usepackage{graphicx}
\usepackage{natbib}
\usepackage{mdframed}
\usepackage{hyperref}
\usepackage{tablefootnote}

\title{The Natural Stories Corpus}
\author{Richard Futrell, Edward Gibson, Hal Tily, Idan Blank, \\Anastasia Vishnevetsky, Steven T. Piantadosi, \\and Evelina Fedorenko}
\date{\today}

\begin{document}
\maketitle

\begin{abstract}
It is now a common practice to compare models of human language processing by predicting participant reactions (such as reading times) to corpora consisting of rich naturalistic linguistic materials.
However, many of the corpora used in these studies are based on naturalistic text
and thus do not contain many of the low-frequency syntactic constructions
that are often required to distinguish processing theories.
Here we describe a new corpus
consisting of English texts edited to contain many low-frequency syntactic constructions
while still sounding fluent to native speakers.
The corpus is annotated with hand-corrected parse trees and includes self-paced reading time data.
Here we give an overview of the content of the corpus and release the data.\footnote{Available from \texttt{http://github.com/languageMIT/naturalstories}. This corpus is distributed under an Attribution-NonCommercial-ShareAlike (CC BY-NC-SA) license, allowing free modification and re-distribution of the corpus so long as derivative work is released under the same terms.}
\end{abstract}

\section{Introduction}
It is becoming a standard practice to evaluate theories of human language processing
by their ability to predict psychometric dependent variables such as per-word reaction time
for standardized corpora of naturalistic text.
Dependent variables that have been collected over fixed corpora
include word fixation time in eyetracking \citep{kennedy2003dundee},
word reaction time in self-paced reading \citep{roark2009deriving,frank2013reading},
BOLD signal in fMRI data \citep{bachrach2009incremental},
and event-related potentials \citep{dambacher2006frequency,frank2015erp}.

The more traditional approach to evaluating psycholinguistic models has been to collect psychometric measures on hand-crafted experimental stimuli designed to tease apart detailed model predictions.
While this approach makes it easy to compare models on their accuracy for specific constructions and phenomena, it is hard to get a sense from experimental results of how models compare on their coverage of a broad range of phenomena.
When it is standard practice to compare model predictions over standardized texts, then it is easier to evaluate coverage.

Although the fixed corpus approach has these advantages, the existing corpora currently used are based on naturally-occurring data,
which is unlikely to include the kinds of sentences which can crucially distinguish between theories.
Many of the most puzzling phenomena in psycholinguistics, and the phenomena which have been used to test models, have only been observed in extremely rare constructions, such as multiply nested preverbal relative clauses \citep{gibson1999memory,grodner2005consequences,vasishth2010shortterm}.
Corpora of naturally-occurring text are unlikely to contain these constructions.
More generally, models of human language comprehension are more likely to make distinct predictions for sentences that cause difficulty for humans, rather than for sentences that are easy to process.
For instance, models of comprehension difficulty based on memory integration cost during parsing \citep{gibson2000dependency,lewis2005activationbased,schuler2010broadcoverage,vanschijndel2013model} will predict effects when the memory spans required for parsing are large,
but most syntactic dependencies in naturally-occurring text are short \citep{temperley2007minimization,liu2008dependency,futrell2015largescale}.
In general, processing difficulty might be rare for naturally-occurring text, because text written and edited in order to be easily understood.

Here we attempt to combine the strength of experimental approaches,
which can test theories using targeted low-frequency structures,
and corpus studies, which provide broad-coverage comparability between models.
We introduce and release a new corpus,
the Natural Stories Corpus,
a series of English narrative texts designed to contain
many low-frequency and psycholinguistically interesting syntactic constructions
while still sounding fluent and coherent.
The texts are annotated with hand-corrected Penn Treebank style phrase structure parses,
and Universal Dependencies parses automatically generated from the phrase structure parses.
We also release self-paced reading time data for all texts,
and word-aligned audio recordings of the texts.
We hope the corpus can form the basis for further annotation and become a standard test set for psycholinguistic models.

\section{Related Work}
Here we survey datasets which are commonly used to test psycholinguistic theories and how they relate to the current release.

Currently the most prominent psycholinguistic corpus for English is the Dundee Corpus \citep{kennedy2003dundee}, which contains 51,501 word tokens in 2,368 sentences from British newspaper editorials, along with eyetracking data from 10 participants.
A dependency parse of the corpus is released in \citet{barrett2015dundee}.
Like in the current work, the eyetracking data in the Dundee corpus is collected for sentences in context and so reflects influences beyond the sentence level.
The corpus has seen wide usage, see for example \citet{demberg2008data,mitchell2010syntactic,frank2011insensitivity,fossum2012sequential,smith2013effect,vanschijndel2015hierarchic,luong2015evaluating}.

The Potsdam Sentence Corpus \citep{kliegl2006tracking} of German provides 1138 words in 144 sentences, with cloze probabilities and eyetracking data for each word.
Like the current corpus, the Potsdam Sentence Corpus was designed to contain varied syntactic structures, rather than being gathered from naturalistic text.
The corpus consists of isolated sentences which do not form a narrative, and during eyetracking data collection the sentences were presented in a random order.
The corpus has been used to evaluate models of sentence processing based on dependency parsing \citep{boston2008parsing,boston2011parallel} and to study effects of predictability on event-related potentials \citep{dambacher2006frequency}.

The MIT Corpus introduced in \citet{bachrach2009incremental} has similar aims to the current work, collecting reading time and fMRI data over sentences designed to contain varied structures.
This dataset consists of four narratives with a total of 2647 tokens; it has been used to evaluate models of incremental prediction in \citet{roark2009deriving}, \citet{wu2010complexity}, and \citet{luong2015evaluating}.

The UCL Corpus \citep{frank2013reading} consists of 361 English sentences drawn from amateur novels, chosen for their ability to be understood out of context, with self-paced reading and eyetracking data.
The goal of the corpus is to provide a sample of typical narrative sentences,
complementary to our goal of providing a corpus with low-frequency constructions.
Unlike the current corpus, the UCL Corpus consists of isolated sentences, so the psychometric data do not reflect effects beyond the sentence level.

Eyetracking corpora for other languages are also available, including the Postdam-Allahabad Hindi Eyetracking Corpus \citep{husain2014integration} and the Beijing Sentence Corpus of Mandarin Chinese \citep{yan2010flexible}.

\section{Corpus Description}

\subsection{Text}

The Natural Stories corpus consists of 10 stories, comprising 10,245 lexical word tokens and 485 sentences in total.
The stories were developed by A.V., E.F., E.G. and S.P. by taking existing publicly available texts and editing them to use many subject- and object-extracted relative clauses, clefts, topicalized structures, extraposed relative clauses, sentential subjects, sentential complements, local structural ambiguity (especially NP/Z ambiguity), idioms, and conjoined clauses with a variety of coherence relations.
The original texts are listed in Table~\ref{tab:story-origins}.

\begin{table}
  \centering
  {\tiny
  \begin{tabular}{|r|l|l|l|}
    \hline
    Story & Title & Source Title & Source Author \\
    \hline
    1 & Boar & The Legend of the Bradford Boar\tablefootnote{http://www.make4fun.com/stories/British-short-story/3917-The-Legend-of-the-Bradford-Boar-by-E-H-Hopkinsona} & E. H. Hopkinson \\
    2 & Aqua & Aqua, or the Water Baby\tablefootnote{http://fullreads.com/literature/aqua-or-the-water-baby/} & Kate Douglas Wiggin \\
    3 & Matchstick & The Little Match-Seller\tablefootnote{http://stenzel.ucdavis.edu/180/anthology/matchgirl.html} & Hans Christian Andersen \\
    4 & King of Birds & The King of the Birds\tablefootnote{http://www.apples4theteacher.com/holidays/bird-day/short-stories/the-king-of-the-birds.html} & Brothers Grimm \\
    5 & Elvis & Elvis Died at the Florida Barber College\tablefootnote{http://www.eastoftheweb.com/short-stories/UBooks/ElvDie.shtml} & Roger Dean Kiser \\
    6 & Mr. Sticky & Mr. Sticky\tablefootnote{http://www.eastoftheweb.com/short-stories/UBooks/MrStic.shtml} & Mo McAuley\\
    7 & High School & Bullies & Sarah Cleaves \\
    8 & Roswell & Roswell UFO incident\tablefootnote{http://en.wikipedia.org/w/index.php?title=Roswell\_UFO\_incident\&oldid=331989741} & Wikipedia\\
    9 & Tulips & Tulip mania\tablefootnote{http://en.wikipedia.org/w/index.php?title=Tulip\_mania\&oldid=329157998} & Wikipedia \\
    10 & Tourette's & Tourette Syndrome Fact Sheet\tablefootnote{http://www.ninds.nih.gov/Disorders/Patient-Caregiver-Education/Fact-Sheets/Tourette-Syndrome-Fact-Sheet} & NINDS \\
    \hline
  \end{tabular}
  }
  \caption{Stories with titles and sources. Footnotes contain URLs for the original text.}
  \label{tab:story-origins}
\end{table}

\begin{figure}
\centering
\includegraphics[scale=.7]{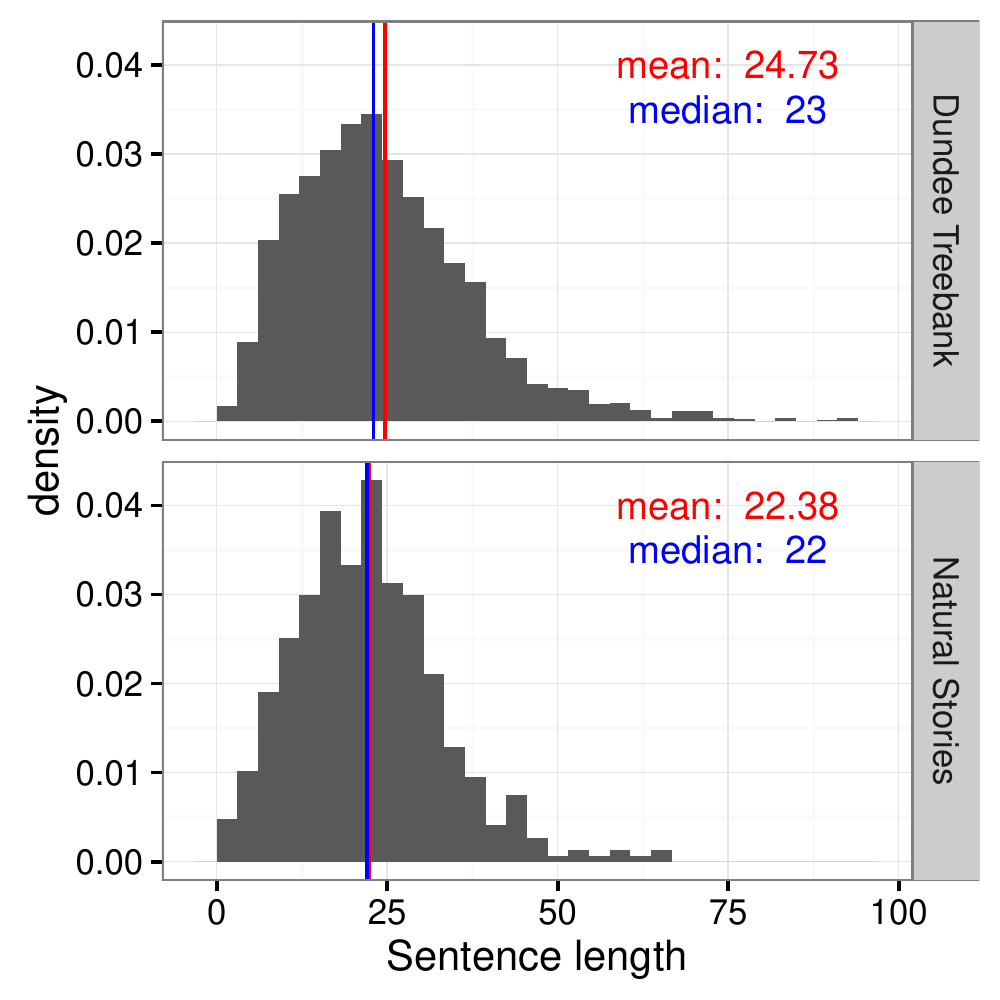}
\caption{Histograms of sentence length (in tokens, including punctuation) in Natural Stories and the Dundee corpus.}
\label{fig:slen}
\end{figure}

The mean number of lexical words per sentence is 21.1, around the same as the Dundee corpus (21.7).
Figure~\ref{fig:slen} shows a histogram of sentence length in Natural Stories as compared to Dundee.
The word and sentence counts for each story are given in Table~\ref{tab:story-counts}.
Each token has a unique code which is referenced throughout the various annotations of the corpus.

\begin{table}
\centering
\begin{tabular}{|r|r|r|}
\hline
Story & \# Words & \# Sentences \\
\hline
1 & 1073 & 57 \\
2 & 990 & 37 \\
3 & 1040 & 55 \\
4 & 1085 & 55 \\
5 & 1013 & 45 \\
6 & 1089 & 64 \\
7 & 999 & 48 \\
8 & 980 & 33 \\
9 & 1038 & 48 \\
10 & 938 & 43 \\
\hline
\end{tabular}
\caption{Summary of stories by length.}
\label{tab:story-counts}
\end{table}



In Figure~\ref{fig:text-sample} we give a sample of text from the corpus (from the first story).
\begin{figure}
  \begin{mdframed}
    If you were to journey to the North of England, you would come to a valley that is surrounded by moors as high as mountains. It is in this valley where you would find the city of Bradford, where once a thousand spinning jennies that hummed and clattered spun wool into money for the long-bearded mill owners. That all mill owners were generally busy as beavers and quite pleased with themselves for being so successful and well off was known to the residents of Bradford, and if you were to go into the city to visit the stately City Hall, you would see there the Crest of the City of Bradford, which those same mill owners created to celebrate their achievements. \\
  \end{mdframed}
  \caption{Sample text from the first story.}
  \label{fig:text-sample}
\end{figure}

\subsection{Parses}

The texts were parsed automatically using the Stanford Parser \citep{klein2003accurate} and hand-corrected.
Trace annotations were added by hand.
We provide the resulting Penn Treebank-style phrase structure parse trees.
We also provide Universal Dependencies parses \citep{nivre2015towards} automatically converted from the corrected parse trees using the Stanford Parser.


\subsection{Self-Paced Reading Data}

\begin{figure}
\centering
\includegraphics[scale=.7]{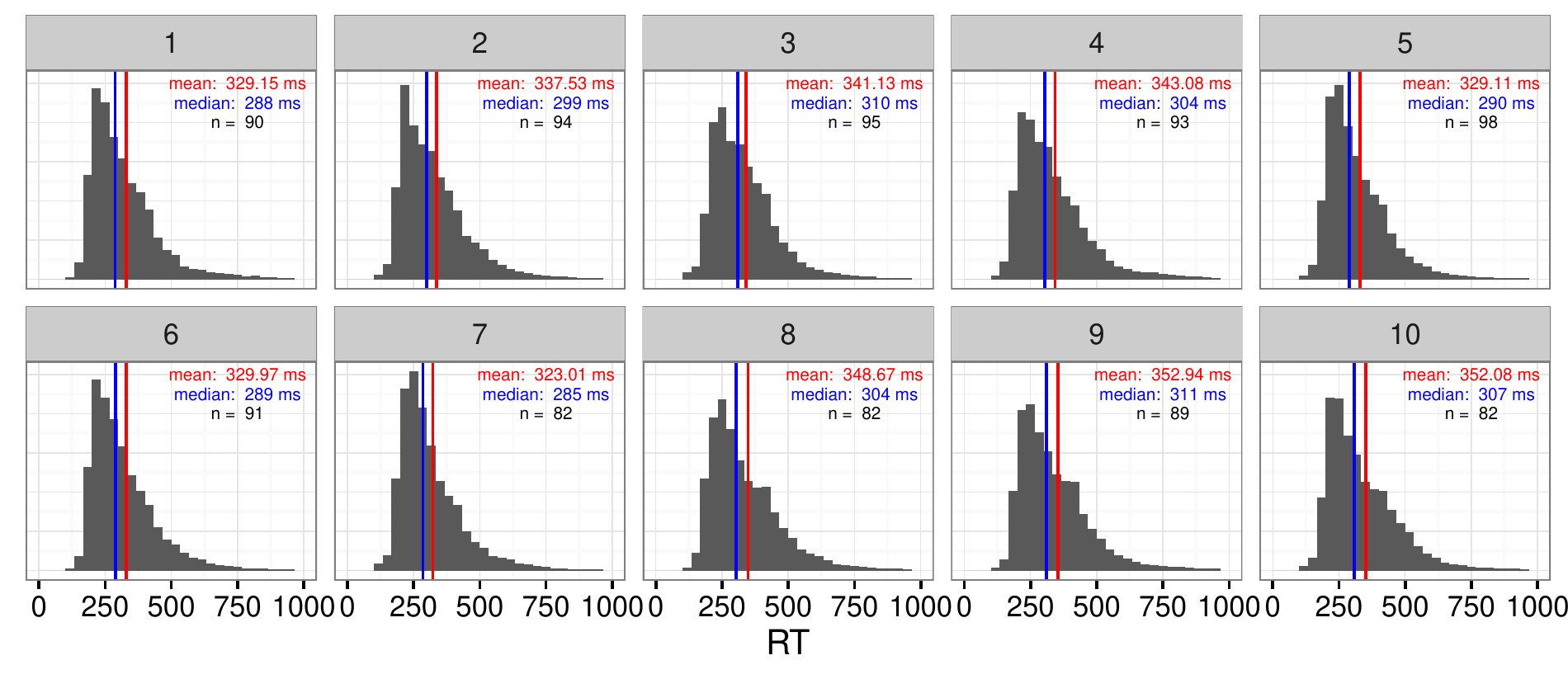}
\caption{Histograms of SPR RTs per story, after data exclusion.}
\label{fig:stories-spr-hist}
\end{figure}

We collected self-paced reading (SPR) data \citep{just1982paradigms} for the stories from 181 native English speakers over Amazon Mechanical Turk.
Text was presented in a dashed moving window display; spaces were masked.
Each participant read 5 stories per HIT.
19 participants read all 10 stories, and 3 participants stopped after one story.
Each story was accompanied by 6 comprehension questions.
We discarded SPR data from a participant's pass through a story if the participant got less than 5 questions correct (89 passes through stories excluded).
We also excluded RTs less than 100 ms or greater than 3000 ms.
Figure~\ref{fig:stories-spr-hist} shows histograms of RTs per story.

\subsubsection{Inter-Subject Correlations}

\begin{figure}
  \includegraphics[scale=.3]{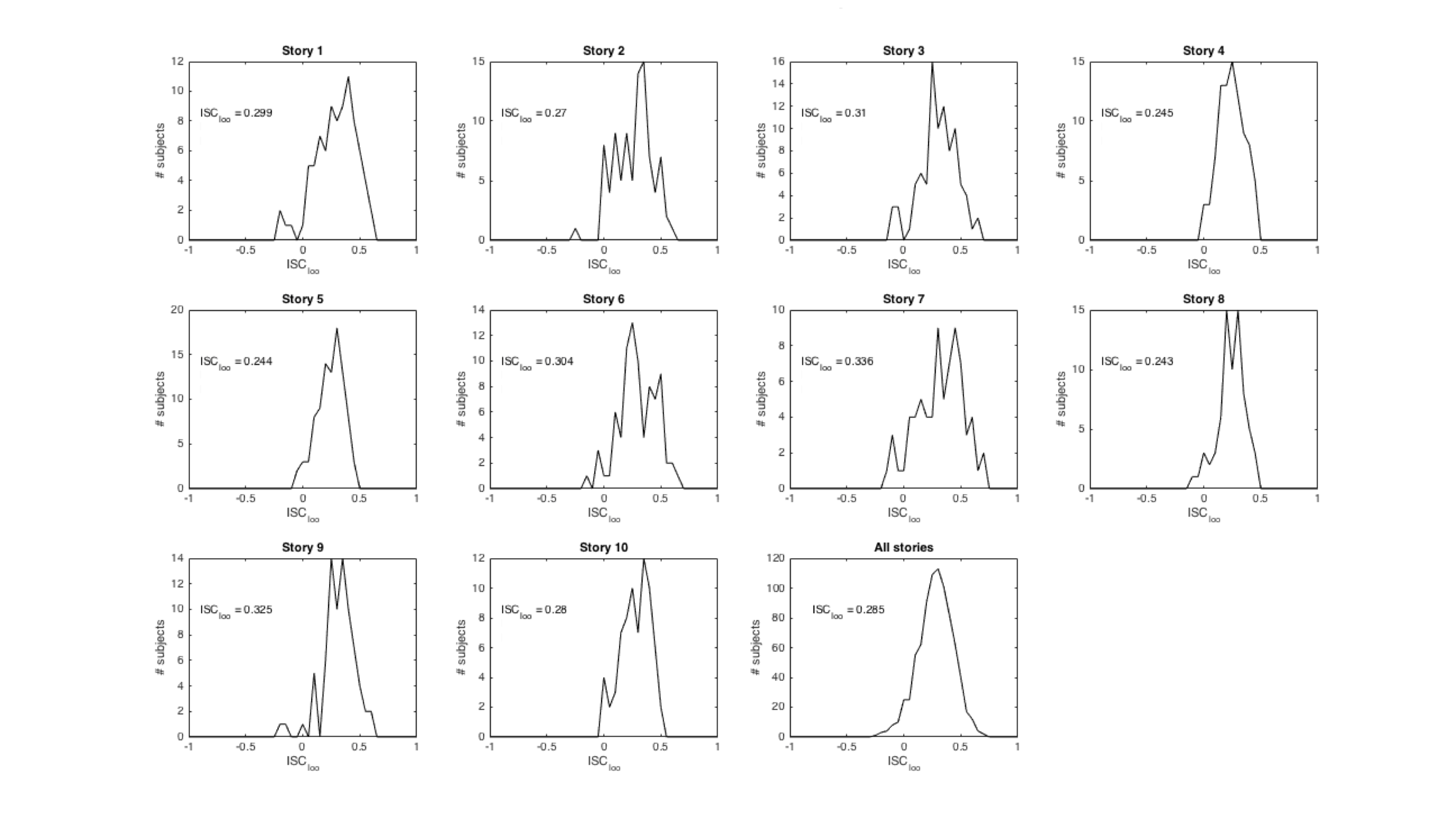}
  \caption{Leave-one-out Inter-Subject Correlations (ISCs) of RTs per story. In the panels, $ISC_{loo}$ gives the average leave-one-out ISC for that story.}
  \label{fig:isc}
\end{figure}

In order to evaluate the reliability of the self-paced reading RTs and their robustness across experimental participants, we analyzed inter-subject correlations (ISCs). For each subject, we correlated the Spearman correlation of that subject's RTs on a story with average RTs from all other subjects on that story. Thus for each story we get one ISC statistic per subject. Figure~\ref{fig:isc} shows histograms of these statistics per story.

\subsubsection{Psycholinguistic Sanity Checks}

In order to validate our RT data, we checked that basic psycholinguistic effects obtain in it.
In particular, we examined whether the well-known effects of frequency, word length, and surprisal \citep{hale2001probabilistic,levy2008expectation,smith2013effect} had an effect on RTs.
To do this, for each of the three predictors log frequency, log trigram probability, and word length, we fit a linear mixed effects regression model with subject and story as random intercepts (models with random slopes did not converge) predicting RT.
Frequency and trigram probabilities were computed from Google Books N-grams, summing over years from 1990 to 2013. (These counts are also released with this dataset.)
The results of the regressions are shown in Table~\ref{tab:surprisal-regression}; we report results from the maximal converging model.
In keeping with well-known effects, increased frequency and probability both lead to faster reading times, and word length leads to slower reading times.

\begin{table}
\centering
\begin{tabular}{|r|r|r|r|r|}
\hline
Predictor & $\hat{\beta}$ & Std. Error & $t$ value & $p$ value \\
\hline
Log Frequency & -2.61 & 0.08 & -32.27 & $<$ 0.001 \\
Trigram Surprisal & -2.19 & 0.09 & -23.90 & $<$ 0.001 \\
Word Length & 4.21 & 0.12 & 35.72 & $<$ 0.001 \\
\hline
\end{tabular}
\caption{Regression coefficients and significance from individual mixed-effects regressions predicting RT for each of the three predictors log frequency, log trigram probability, and word length.}
\label{tab:surprisal-regression}
\end{table}

\subsection{Syntactic Constructions}
\label{sec:construction-survey}
Here we give an overview of the low-frequency or marked syntactic constructions which occur in the stories.
We coded sentences in the Natural Stories corpus for presence of a number of marked constructions, and also coded 200 randomly selected sentences from the Dundee corpus for the same features.
The features coded are listed and explained in Appendix~\ref{sec:features-coded}.
Figure~\ref{fig:ns-dundee} shows the rates of these marked constructions per sentence in the two corpora.
From the figure, we see that the natural stories have especially high rates of nonlocal VP conjunction, nonrestrictive SRCs, idioms, adjective conjunction, noncanonical ORCs, local NP/S ambiguities, and it-clefts.

\begin{figure}
  \centering
  \includegraphics[scale=.5]{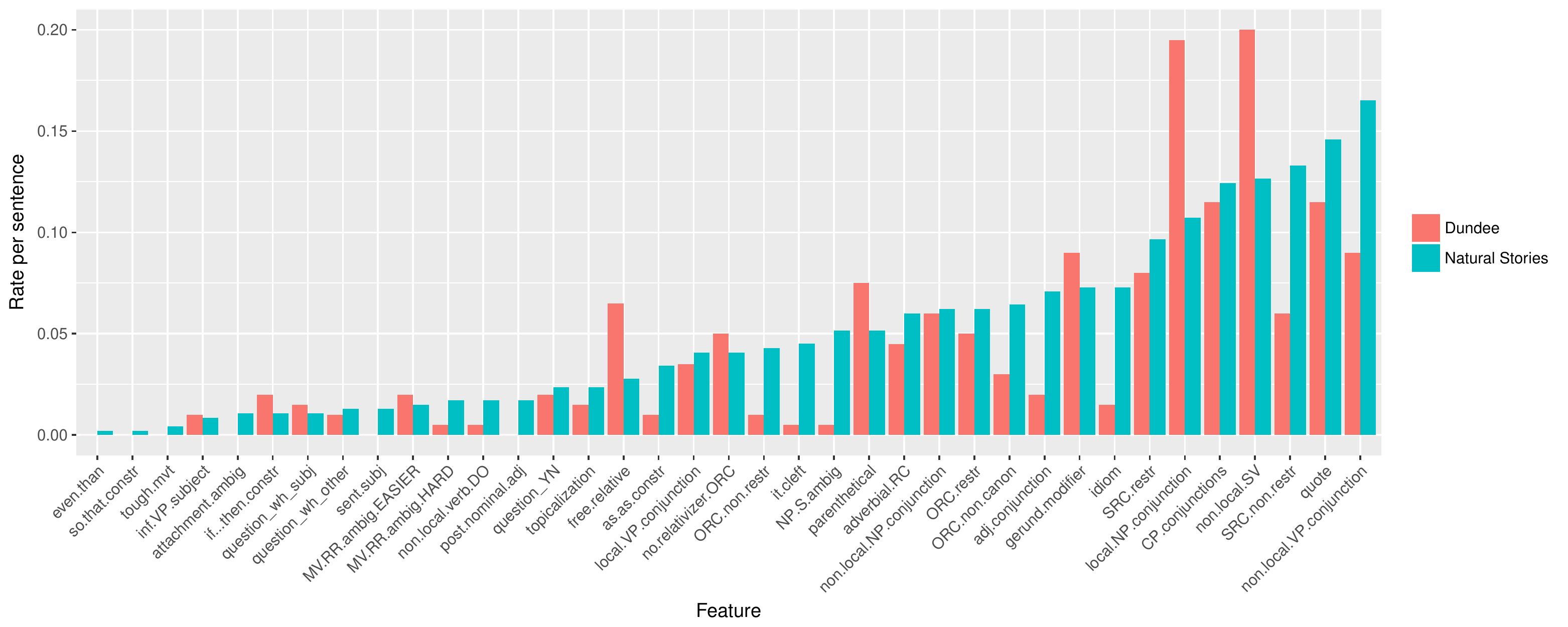}
  \caption{Rates of marked constructions in the Natural Stories corpus and in 200 randomly sampled sentences from the Dundee corpus.}
  \label{fig:ns-dundee}
\end{figure}

\section{Conclusion}
We have described a new psycholinguistic corpus of English,
consisting of edited naturalistic text designed to contain many rare or hard-to-process constructions
while still sounding fluent.
We believe this corpus will provide an important part of a suite of test sets for psycholinguistic models,
exposing their behavior in uncommon constructions
in a way that fully naturalistic corpora cannot.
We also hope that the corpus as described here forms the basis for further data collection and annotation.

\section*{Acknowledgments}

This work was supported by NSF DDRI grant \#1551543 to R.F., NSF grants \#0844472 and \#1534318 to E.G., and NIH career development award HD057522 to E.F. The authors thank the following individuals: Laura Stearns for hand-checking and correcting the parses, Suniyya Waraich for help with syntactic coding, Cory Shain and Marten van Schijndel for hand-annotating the parses, and Kyle Mahowald for help with initial exploratory analyses of the SPR data.
The authors also thank Nancy Kanwisher for recording half of the stories (the other half was recorded by E.G.), Wade Shen for providing initial alignment between the audio files and the texts, and Jeanne Gallee for hand-correcting the alignment.

\bibliography{everything}{}
\bibliographystyle{apa}

\appendix
\section{Features coded for Section~\ref{sec:construction-survey}}
\label{sec:features-coded}
The features coded are:
\begin{itemize}
\item{Local/nonlocal VP conjunction}: Conjunction of VPs in which the head verbs are adjacent (local) or not adjacent (nonlocal)
\item{Local/nonlocal NP conjunction}: Conjunction of VPs in which the head nouns are adjacent (local) or not adjacent (nonlocal). 
\item{Sentential conjunction}: Conjunction of sentences.
\item{CP conjunction}: Conjunction of CPs with explicit quantifiers.

\item{Restrictive/nonrestrictive SRC}: Subject-extracted relative clauses with either restrictive or nonrestrictive semantics
\item{Restrictive/nonrestrictive ORC}: Object-extracted relative clauses with either restrictive or nonrestrictive semantics
\item{No-relativizer ORC}: An object-extracted relative clause without an explicit relativizer, e.g. \emph{The man I know}
\item{Noncanonical ORC}: An object-extracted relative clause where the subject is not a pronoun.
\item{Adverbial relative clause}: An relative clause with an extracted adverbial, e.g. \emph{the valley \underline{where you would find the city of Bradford}}.
\item{Free relative clause}

\item{NP/S ambiguity}: A local ambiguity where it is unclear whether a clause is an NP or the subject of a sentence. For example, \emph{I know \underline{Bob} is a doctor}.
\item{Main Verb/Reduced Relative ambiguity (easy/hard)}: A local ambiguity between a main verb and a reduced relative clause. For example, \emph{The horse \underline{raced} past the barn fell}.
\item{PP attachment ambiguity}

\item{Nonlocal SV}: The appearance of any material between a verb and the head of its subject.
\item{Nonlocal Verb/DO}: The appearance of any material between a verb and its direct object.
\item{Gerund modifiers}
\item{Sentential subject}
\item{Parentheticals}
\item{\emph{Tough} movement}
\item{Postnominal adjectives}
\item{Topicalization}
\item{even...than construction}
\item{if...then construction}
\item{as...as construction}
\item{Yes-No Question}
\item{Question with \emph{wh} subject}
\item{Question with other \emph{wh} word}
\item{Idiom}: Any idiomatic expression, such as \emph{busy as beavers}.
\item{Quotation}: Any directly-reported speech
\end{itemize}

Coding was performed by Suniyya Waraich, Edward Gibson, and Richard Futrell.

\end{document}